\documentclass[letterpaper, 10 pt, conference]{ieeeconf}  

\IEEEoverridecommandlockouts                 

\usepackage{amsmath} 
\usepackage{amssymb}  
\usepackage{algpseudocode}
\usepackage{graphicx}
\usepackage{physics,bm,bbm}

\usepackage{multirow}
\usepackage{cuted}
\usepackage{cite}

\usepackage{booktabs}
\usepackage{colortbl}
\usepackage{multicol}

\usepackage{makecell}
\usepackage{subcaption}
\let\labelindent\relax
\usepackage{enumitem}
\usepackage{pifont}
\newcommand{\xmark}{\ding{55}}
\newcommand{\cmark}{\ding{51}}
\usepackage[ruled,vlined]{algorithm2e}
\usepackage{caption}
\usepackage{subcaption} 
\usepackage{dblfloatfix} 
\usepackage{adjustbox} 

\usepackage{xcolor}

\definecolor{nhred}{HTML}{D62727}

\title{\LARGE \bf
Learning to Design Soft Hands using Reward Models
}

\author{
Xueqian Bai$^{1}$,
Nicklas Hansen$^{1}$,
Adabhav Singh$^{1}$,
Michael T. Tolley$^{1}$,
Yan Duan$^{2}$,
Pieter Abbeel$^{2}$,\\
Xiaolong Wang$^{1}$,
Sha Yi$^{1}$%
}

\begin{document}

\twocolumn[{%
\renewcommand\twocolumn[1][]{#1}%
\maketitle
\vspace{-1em}
\begin{center}
\vspace{-15pt}
  $^{1}$UC San Diego, $^{2}$Amazon FAR (Frontier AI \& Robotics)
\end{center}
\includegraphics[width=0.95\linewidth]{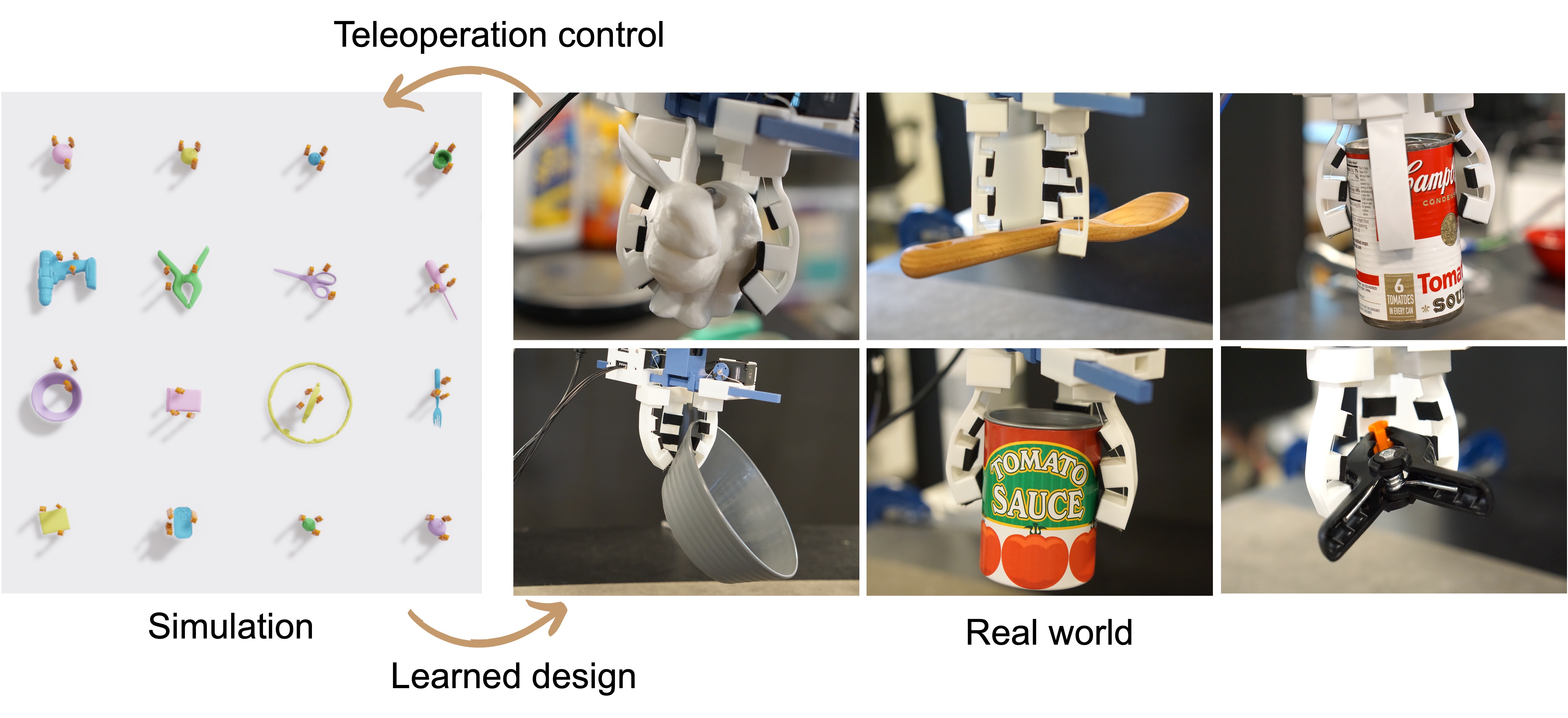}

\captionof{figure}{We present a Cross-Entropy Method (CEM) with reward model (\textbf{CEM-RM}) framework that optimizes block-wise, finger-wise, and tendon-routing design distributions of a soft robotic hand using pre-collected teleoperation data. Hardware experiments demonstrate that CEM-RM achieves effective design optimization with significantly fewer samples than pure optimization, enabling robust grasping of challenging objects. \vspace{1em}}
\label{fig:teaser}
}]
\thispagestyle{empty}
\pagestyle{empty}


\begin{abstract}
Soft robotic hands promise to provide compliant and safe interaction with objects and environments. However, designing soft hands to be both compliant and functional across diverse use cases remains challenging. Although
co-design of hardware and control better couples morphology to behavior\cite{morph}, the resulting search space is high-dimensional, and even simulation-based evaluation is computationally expensive. In this paper, we propose a Cross-Entropy Method with Reward Model (CEM-RM) framework that efficiently optimizes tendon-driven soft robotic hands based on teleoperation control policy, reducing design evaluations by more than half compared to pure optimization while learning a distribution of optimized hand designs from pre-collected teleoperation data. We derive a design space for a soft robotic hand composed of flexural soft fingers and implement parallelized training in simulation. The optimized hands are then 3D-printed and deployed in the real world using both teleoperation data and real-time teleoperation. Experiments in both simulation and hardware demonstrate that our optimized design significantly outperforms baseline hands in grasping success rates across a diverse set of challenging objects.

\end{abstract}


\section{Introduction}
Recent progress in teleoperation~\cite{aloha, yang2024ace} and rapid fabrication technologies~\cite{khatib2016ocean} have created powerful opportunities for developing and evaluating novel robotic hand designs. Many of these are done with rigid robotic hands, which are straightforward to model and control~\cite{xin2025analyzingkeyobjectiveshumantorobot}. However, rigid robotic hands remain costly as they require small and accurate motors, and tend to be fragile when executing contact-rich interactions in a tabletop setting. The 3D printers have become accessible and greatly enabled rapid prototyping of alternative hand morphologies, including soft hands. Soft hands are inherently safe to work around humans, objects, and environments. Soft fingers actuated with tendons~\cite{odhner2014compliant} are especially easy to make: they can be directly 3D-printed in a single piece~\cite{zhai2023desktop, yi2025codesign} with integrated flexure joints. This allows us to quickly explore a broad design space by validating the hardware design in the real world.

Advances in computational methods and machine learning are transforming how robotic design and control can be optimized. Gradient- and planning-based methods have been explored to optimize the geometry of robots\cite{schulz2017interactive, liu2018optimal}. However, these methods typically require extensive domain expertise. Machine learning offers a data-driven alternative~\cite{ha2021fit2form, xu2024dynamics}, yet design problems are still challenging because they lack direct supervision: engineers often rely on intuition or costly trial-and-error iterations. This is particularly limiting when designs must generalize across different use cases. Co-design of both control and design has been shown to effectively connect designs with their use cases~\cite{xu2021end, yi2025codesign}. However, design and control by themselves are hard problems with high dimensionality. The joint co-design space can quickly become intractable.

In this work, we present a framework for learning to design soft robotic hands using a reward model. Our approach uses control demonstrations from teleoperation collected on real objects and transfers them into simulation. We sample a wide range of hand design parameters and simulate the corresponding grasping outcomes. To explore the design space efficiently, we employ the Cross-Entropy Method (CEM), which maintains and iteratively refines a sampling distribution over design parameters by resampling from the top-performing candidates. However, this process can still be slow and costly. Thus, we use the simulation grasping outcomes as ground-truth rewards to train a neural reward model based on a multilayer perceptron (MLP). By combining the reward model with CEM, we significantly accelerate the design search: the reward model reduces the required simulation budget by half while maintaining its fidelity. We then validate the optimized designs in both simulation and hardware by fabricating hands via 3D printing. Results show that our learned designs outperform a uniform baseline, achieving higher grasp success rates.

Overall, our approach demonstrates how to efficiently explore large design space using sampling and reward models. More broadly, it offers an option for generating optimal robot designs given an existing control policy, without requiring domain-specific analytic models or excessive simulation trials.

\section{Related Work}
Robotic hand design faces challenges from high-dimensional parameter spaces and the complexity of optimizing control policies. Recent advances in machine learning and world models have improved sample efficiency, while teleoperation offers reliable control demonstrations that bypass the need to learn policies entirely from scratch.

\textbf{Computational Design and Co-Design.} Computational methods use optimization algorithms, simulation, and data analysis to enhance the design process. Prior work has explored planning~\cite{kodnongbua2023computational} and gradient-based optimization over geometry and topology~\cite{chen2018topology}. However, such approaches often fail to generalize across tasks and struggle with discontinuous contacts or fabrication constraints that make gradients intractable. Bi-level formulations jointly optimizing design and control~\cite{ sliacka2023co,zhao2020robogrammar} alleviate this, while simulators~\cite{chen2025physics} enable end-to-end co-design but still remain challenging with non-linearity and sensitive to initialization~\cite{xu2021end, georgiev2024pwm}. Sampling-based frameworks instead trade computational cost for broader exploration: Genetic Algorithms~\cite{kulz2024optimizing} and Covariance Matrix Adaptation - Evolution
Strategy (CMA-ES)~\cite{cma-es} evaluate large populations in discrete or continuous spaces; Bayesian Optimization balances exploration and exploitation in low-dimensional spaces~\cite{ryu2025machine}; and Reinforcement Learning (RL) coupled with evolutionary methods discovers unintuitive morphologies~\cite{softzoo, morph, crawlers}. 

\textbf{Tendon-driven Robotic Hands.} Tendons are a popular way of actuating robotics hands~\cite{odhner2014compliant}. Compared to direct-drive hands such as LEAP~\cite{shaw2023leap}, tendon-driven hands~\cite{gilday2025embodied, zorin2025ruka, toshimitsu2023getting, kim2025exo} take inspiration from biological hands to control joints with tendons driven by actuators positioned away from fingers. 
Some soft designs encode joints directly into monolithic bodies~\cite{zhai2023desktop, mohammadi2020practical} or exploit variable-stiffness materials for adaptable compliance~\cite{kim2019continuously}. However, designing a tendon-driven soft robot body still relies on manual tuning and the exploration of geometric parameters.


\textbf{Optimization with Function Approximation.} Data-driven approaches to optimization such as Reinforcement Learning (RL) have flourished in the era of neural networks, and a variety of broadly applicable RL algorithms have been proposed. These algorithms typically have an \emph{``actor-critic''} structure in which a \emph{critic} approximates the (usually temporally extended) problem landscape, and an \emph{actor} concurrently learns to maximize expected return as estimated by the critic \cite{mnih2013playing, Lillicrap2015ContinuousCW, schulman2017proximal, Haarnoja2018SoftAA}. A key benefit of such methods is their ability to model high-dimensional and often highly nonlinear problem landscapes. Most recently, researchers find that such algorithms can be simplified by replacing the actor with a non-parametric function such as the Cross-Entropy Method (CEM) \cite{RUBINSTEIN199789, kalashnikov2018scalable, lowrey2018plan, Bhardwaj2021BlendingM, sikchi2022learning, Hansen2022tdmpc}, or the critic with Monte Carlo rollouts \cite{williams1992Reinforce, guo2025deepseek}. However, the aforementioned methods were all designed for sequential decision-making rather than the (single-step) continuous bandit problem that we consider in this work. We propose a simple and computationally efficient method for design optimization that uses CEM as actor, and a combination of single-step Monte Carlo rollouts and a learned reward model as critic.


\section{Methods}
Soft robot hands can be designed with tendon-driven compliant fingers for lightweight, dexterous manipulation~\cite{odhner2014compliant}. We define a design space that parameterizes key structural variables, including finger length and thickness, tendon routing, and finger position and orientation, which influence contact forces and grasp stability. We collect teleoperation control data in both simulation and the real workspace to serve as a base for design optimization. Building on this dataset, we introduce the Cross-Entropy Method with reward model (CEM-RM) framework to enable efficient evaluation of design candidates, while CEM iteratively updates the design distribution to converge toward effective soft hand configurations.

\begin{figure*}[thpb]
    \centering
    \begin{subfigure}{0.5\textwidth}
    \centering
\includegraphics[width=\textwidth]{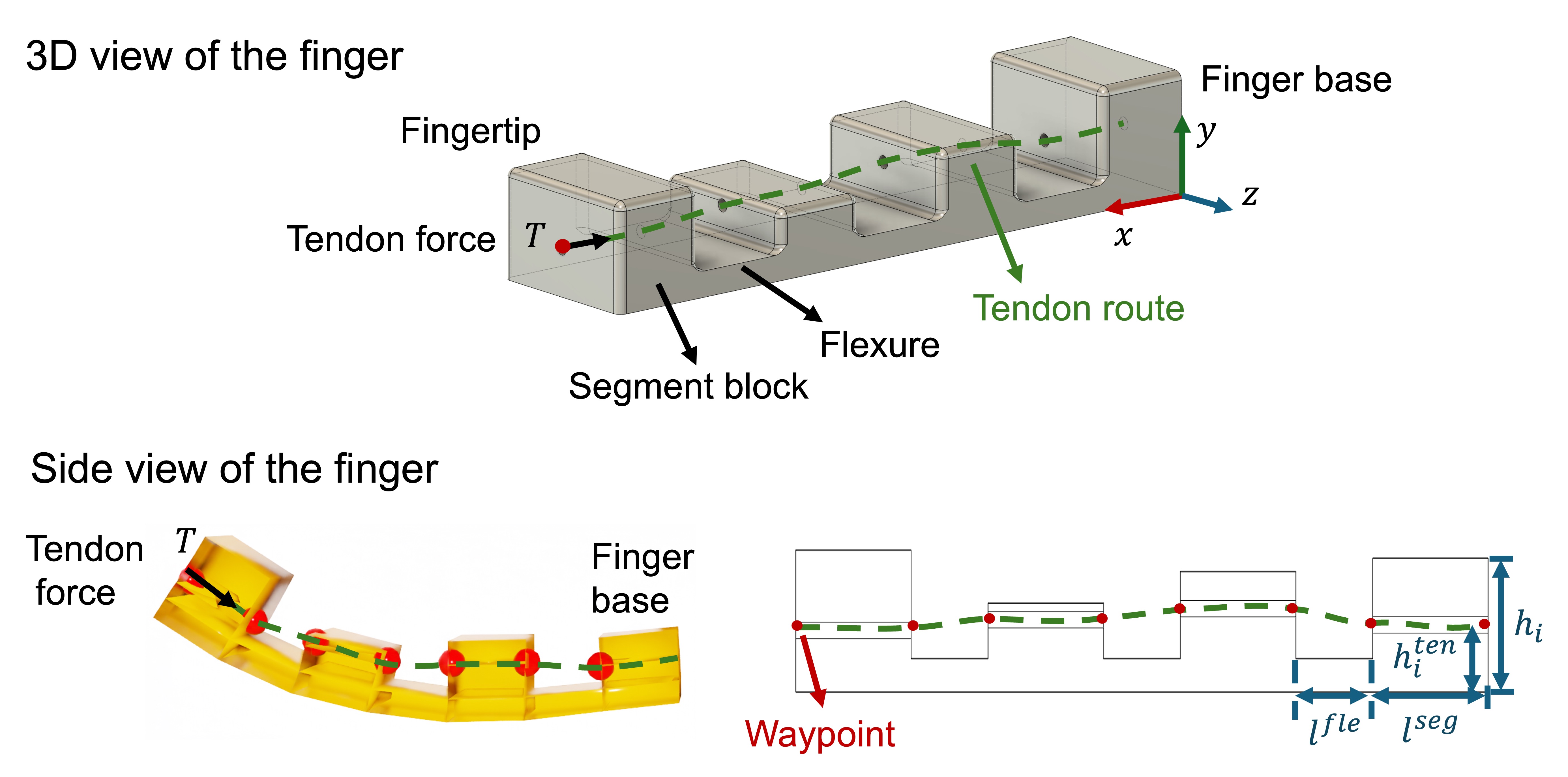}
    \caption{}
    \label{fig:tendon_route}
    \end{subfigure}
    \hspace{0.8cm}
    \begin{subfigure}{0.25\textwidth}
    \centering
    \includegraphics[width=\textwidth]{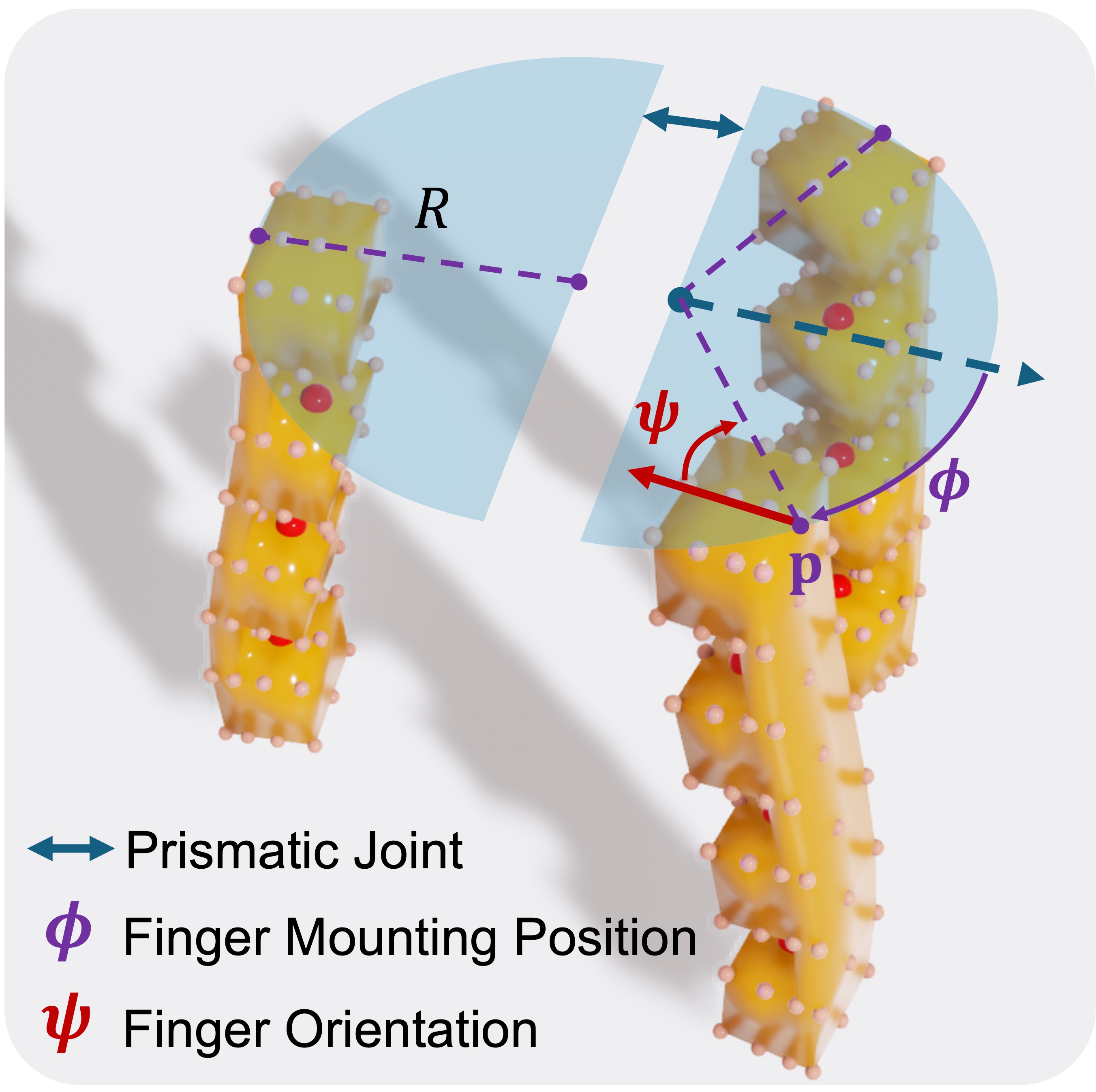}
    \caption{}
    \label{fig:finger_position}
    \end{subfigure}
    \caption{\textbf{Soft robot hand design space.} (a) 3D and side views of a finger, with the fingertip on the left and the base on the right. The tendon force $T$ is applied along the green tendon route toward the base, passing through red waypoints. Optimized parameters include segment and flexure lengths, tendon waypoint distribution, and segment thickness. (b) A three-finger soft hand (base on top, fingers pointing down), where finger orientation and mounting position are also design parameters.}
    \label{fig:param}
    \vspace{-2em}
\end{figure*}
\subsection{Soft Hand Design Space}
We developed a parallel tendon-driven soft-hand simulator in Nvidia Warp \cite{warp2022, newton}. Each finger is actuated by a tendon routed through the one-piece finger mesh via a set of waypoints. We simulated the motion of this actuated finger using the Finite-Element Method (FEM)\cite{yi2025codesign} to model soft-hand deformation. To ensure stable and efficient simulation, we fixed the material stiffness of each finger to a moderate Young’s modulus of 2.0 MPa in our simulation, and optimized the remaining hand parameters.

The optimization variables include finger length, tendon routing, segment-block thickness, and each finger’s mounting position and orientation (see Fig.~\ref{fig:param}). 
Finger-length–related variables are denoted by $l$, where $l^{fle}$ represents the flexure length and $l^{seg}$ the segment-block length. Both variables range from 6 to 18 mm, which is sufficient for the fingers to handle most object sizes while ensuring stable simulation. Tendon-route and block-thickness variables are denoted by $h$. Tendon waypoints are positioned on both sides of each segment block; for segment block $i$, we design both its thickness $h_i$ and the tendon-waypoint height $h_i^{ten}$. $h_i$ ranges from 4 to 18 mm, ensuring that the segment block is taller than the flexure and provides sufficient frictional support. $h_i^{ten}$ also lies within this range but is constrained to remain below the segment height.

We formulate the mounting positions using two half-circles, with the prismatic joint located between them. The thumb is mounted on one side of the joint, while the index and middle fingers are mounted on the other side. We denote each finger’s mounting orientation by $\psi$ and mounting position by $\mathbf{p}$ (corresponding to the mounting angle $\phi$):
\begin{align}
\mathbf{p}=R
\begin{bmatrix}
cos\phi\\
\sin\phi
\end{bmatrix}
\end{align}
where $R$ is the finger-mounting half-circle radius. Fingers can be mounted within their corresponding half-circle, while $\psi \in (-\pi/4, \pi/4)$ allows inward-oriented forces.

\begin{figure*}[thpb]
    \centering
    \includegraphics[width=0.95\linewidth]{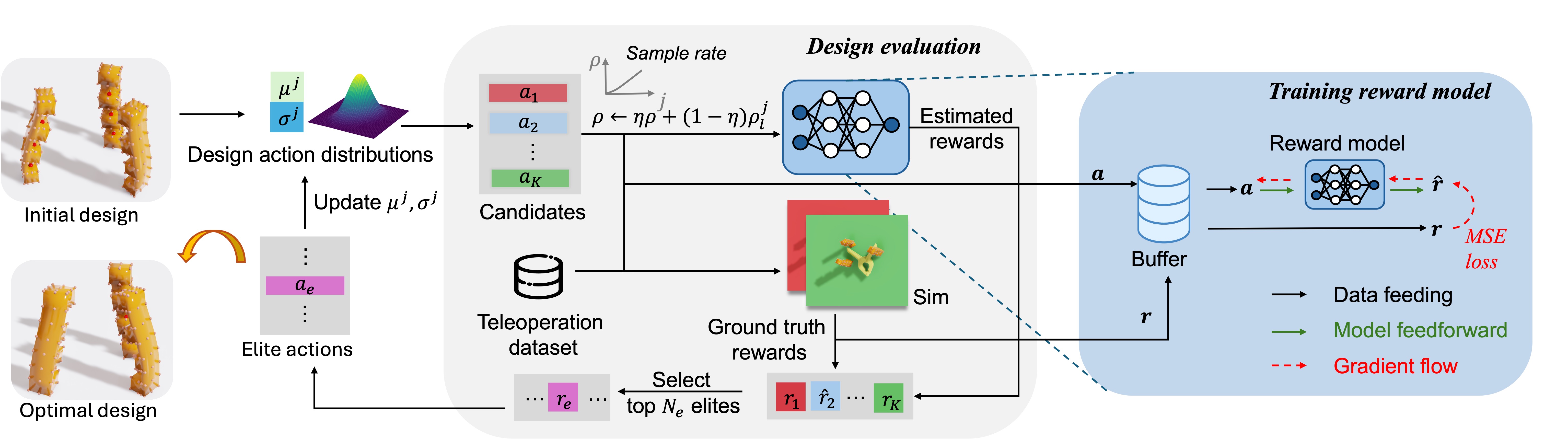}
    \caption{\textbf{System overview.} We first collected multiple teleoperation control datasets for each object, which are randomly sampled during optimization. The design action distribution is optimized in the CEM loop, with evaluations from both simulation and a co-trained reward model (the proportion of reward model evaluations increases smoothly during training). The action distribution ultimately converges to the optimal soft hand design.}
    \label{fig:pipeline}
    \vspace{-1.0em}
\end{figure*}

\subsection{Cross-Entropy Method with Reward Models}
We define our soft hand design optimization as a \textit{continuous bandit problem}. We assume the design variables follow a Gaussian distribution $\mathcal{N}(\mu,\sigma^2I)$ and use the Cross-Entropy Method (CEM) to iteratively update this distribution toward the optimum. As illustrated in Fig.~\ref{fig:pipeline}, starting from a uniform initial design, we feed forward design actions during training while iteratively optimizing the CEM action distribution.
We jointly optimize the soft hand design across all objects in our dataset. We denote $\Delta q$ as the displacement of the object’s ending pose with respect to the soft hand’s wrist, and $\Delta q_y$ as its component along the positive axis pointing upwards. $\mathcal I\in \{0,1\}$, where $\mathcal I=1$ if the object collides with the ground, and 0 otherwise. The objective for an object set $\mathcal{O}$ is formulated as (all variables are conditioned on $l$, $h$, $\mathbf{p}$, and $\mathbf{\psi}$, but omitted for simplicity):
\begin{align}
    \mathcal{R}_{opt}(l,h,\mathbf{p},\psi)
= w_1 \sum_{o \in \mathcal{O}} \lVert \Delta q \rVert +
w_2 \sum_{o \in \mathcal{O}} \lvert \min(\Delta q_y, 0) \rvert - \mathcal I
\label{eq:Objective}
\end{align}
The reward of each valid design (no penetration among fingers) is computed based on this objective. The reward is set to zero if the design is invalid or if there is collision between the object and the ground.


\vspace{-0.8em}
\begin{algorithm}[htbp]
\footnotesize
\KwIn{Multiple teleop motion $\mathcal D_{teleop}$ for each object, initial design $\mathbf s_0$, learned network parameters $\theta$, initial parameters $\mu^0,\sigma^0$ of CEM, replay buffer $\mathcal B$ of (action $a$, reward $r$) from simulation, eval rate $\rho$ from reward model }
\KwOut{Optimized final one-step design state $\mathbf s^*=\mathbf s_0+a^*$}
\BlankLine
\For{iteration $j\leftarrow1$ \KwTo $J$}{
  
  \If{$\rho$ is None}{
    $\rho = \rho_l^j$
  }
  \Else{
    $\rho \leftarrow \eta\rho + (1-\eta)\rho_l^j$
  }
  
  Select population indices of size $\operatorname{round}(\rho \cdot K)$: $\mathcal{I}\sim\mathcal{U}(0, K)$
  
  Collect a set of actions: $\{a_k\}\sim\mathcal{N}(\mu^{j-1}, (\sigma^{j-1})^2I)$

  \For{$k\leftarrow1$ \KwTo $K$}{
  \If{$k\in\mathcal{I}$}{$r_k = Q(a_k|\theta)$}
  \Else{
  Evaluate mean reward on all objects:
  
  $s_k = s_0 + a_k$, $r_k=\mathrm{Sim}(s_k, D_{teleop})$
  
  Store action-reward mapping $(a_k, r_k)$ in $\mathcal B$
  }
  }
  Select a subset of elites $E$ (top $N_e$ samples)
  
  $\mu^j\leftarrow \frac{1}{N_e}\sum\limits_{x\in E}(x)$, $ \sigma^j \leftarrow \sqrt{\frac{1}{N_e}\sum\limits_{x\in E}\lVert x - \mu^j \rVert^2_2} $

  \If{buffer size $>$ batch size $N_B$}{
  Sample a random batch of $N_B$ mappings $(a_i, r_i)$ from buffer $\mathcal{B}$
  
  Update Q by minimizing the loss: \\$L=\frac{1}{N_B}\sum_i(r_i-Q(a_i|\theta))^2$

  $\theta \leftarrow \theta - \alpha\nabla_{\theta}L$
  }
}
\Return $\mathbf{s}^*$\;

  \caption{CEM with reward model ($\mathcal D_{teleop},\, \mathbf s_0$)}
  \label{alg:cem}
\end{algorithm}
\vspace{-0.8em}
In Algorithm~\ref{alg:cem}, we start with multiple teleoperation control data collected beforehand (randomly sampled during design evaluation), an initial design $\mathbf{s_0}$, CEM parameters $\mu^0$ and $\sigma^0$, learned reward model parameters $\theta$, and buffer $\mathcal{B}$ of (action $a$, reward $r$) from simulation. In each optimization iteration $j$, we first sample design actions $a$ from the current CEM distribution $\mathcal{N}(\mu^j,(\sigma^j)^2I)$. Each design action candidate is then evaluated on all objects in parallel, using simulation and the reward model with proportions $1-\rho$ and $\rho$, respectively. The evaluation rate $\rho$ smoothly increases as:
\begin{align}
    \rho\leftarrow \eta\rho+(1-\eta)\rho_l^j
    \label{eq:rho}
\end{align}
where $\rho_l^j=\rho_{min}+\frac{1}{N}(\rho_{max}-\rho_{min})\cdot j$ denotes the linear rate at iteration $j$. The top $N_e$ elite design actions are selected to update the CEM distribution. All ground-truth action-reward mappings from the simulation are stored in a buffer, from which mini-batches are sampled to train the reward model by minimizing the Mean Squared Error (MSE) between model's estimated rewards and the ground-truth rewards:
\begin{align}
L=\frac{1}{N_B}\sum_i(r_i-Q(a_i|\theta))^2
\end{align}
After iterative optimization of CEM with reward model, the algorithm returns the optimal soft hand design $\mathbf{s}^*$ for the object set.
\begin{figure}[tbp]
    \centering
    \includegraphics[width=0.85\linewidth]{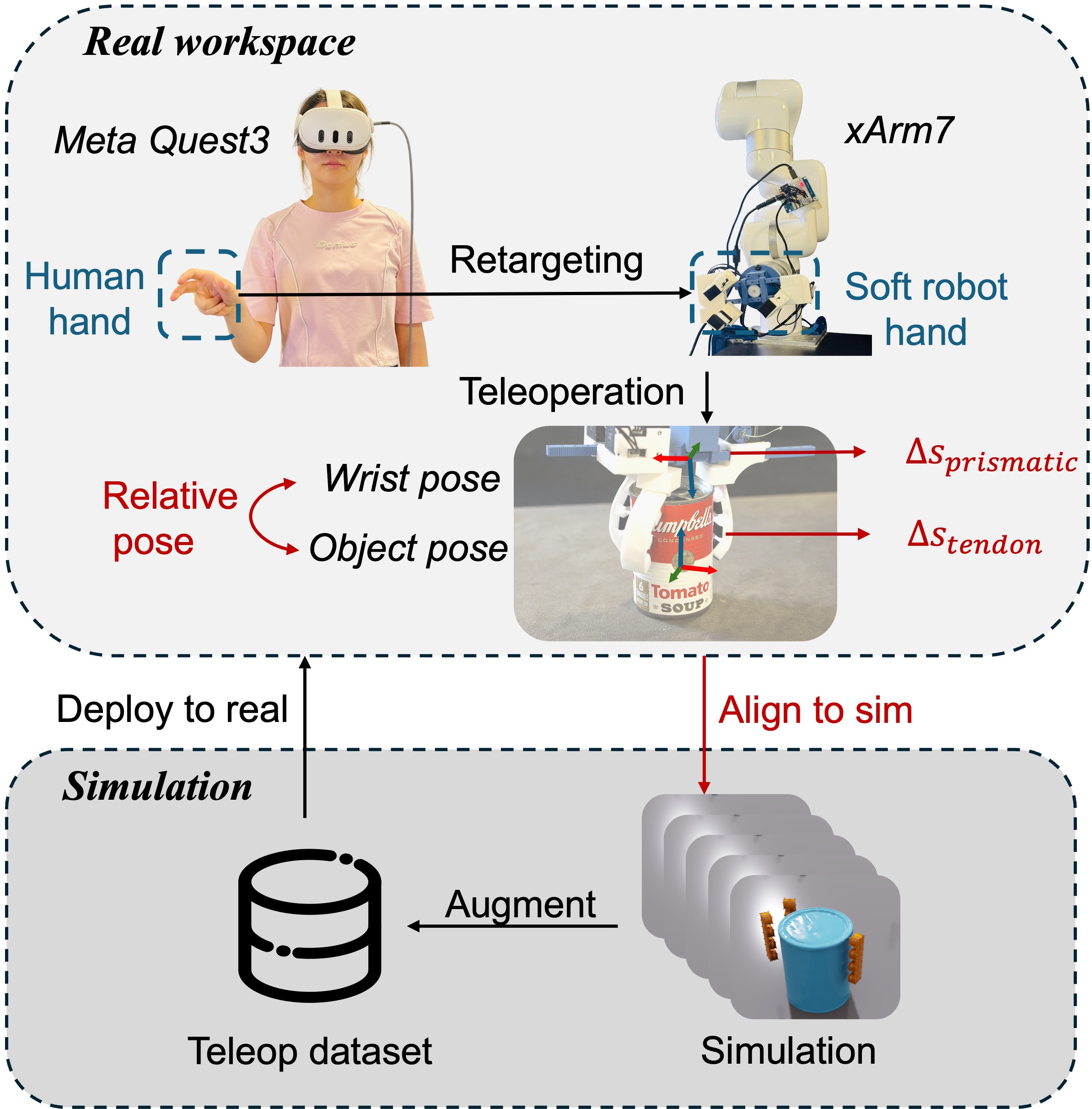}
    \caption{\textbf{Teleoperation data collection.} Human hand poses captured by the Meta Quest 3 are converted into soft-hand motion control commands for real-time teleoperation. The grasping pose, prismatic joint displacement, and tendon motion are collected and augmented in simulation. }
    \label{fig:teleop}
    \vspace{-1.8em}
\end{figure}
\subsection{Teleoperation Data Collection}
To efficiently deploy soft-hand designs with high-quality control motions, we collected teleoperation data using a uniform baseline design in both the real world and simulation, which then serves as a base for the design optimization process (Fig.~\ref{fig:teleop}).
\subsubsection{Teleoperation and Retargeting}
In the real hardware workspace, we use a Virtual Reality (VR) device (Meta Quest 3, Meta Platforms, Inc.) to teleoperate a robot arm (xArm, UFactory USA), the prismatic joint of the soft hand, and the tendon-driven finger motions. The end effector is controlled through delta-pose mapping: the change in the human wrist pose between consecutive time steps is applied directly to the end effector of the robot arm. We detect pose jumps by checking whether two successive transformation matrices remain within a predefined tolerance. For smoother transitions, we apply spherical linear interpolation (SLERP) to quaternions and linear interpolation to translations.
For the soft-hand actuation, the distance between the human thumb and index fingertips is projected to the prismatic joint, mapping directly to the displacement of the prismatic motor (corresponding to prismatic-joint extension). Also, human finger bending is retargeted to the tendon displacement of the soft fingers. We measure bending degree using $ cos\theta = \frac{\vec{v_1}\cdot\vec{v_2}}{\lVert\vec{v_1}\rVert \lVert\vec{v_2}\rVert}$, where $\vec{v_1} = \vec{p}_{fingertip} - \vec{p}_{distal}$, $\vec{v_2} = \vec{p}_{distal}- \vec{p}_{proximal}$, $\vec{p}$ denotes the 3D hand keypoint position in the Quest 3 world frame and $\lVert\cdot\rVert$ is the Euclidean norm.
\vspace{-0.3em}
\subsubsection{Real-to-Sim and Simulation Data Collection}
We collect three types of control signals during teleoperation: (a) the 6D end-effector grasping pose relative to the object pose, (b) the prismatic-joint displacement, and (c) tendon displacements of the soft fingers. The linear displacement along the motor pulley $\Delta s $ is computed as:
$\Delta s = r_{pulley}\Delta q$, where $\Delta q$ denotes the motor joint angle change, and $r_{pulley}$ is the pulley radius.
Then we align the teleoperation data to the simulation by executing velocity control for both the prismatic joint and lift-up motions. This avoids finger–object penetration: over a short horizon we first apply a constant positive acceleration finger mesh, then an equal-magnitude negative acceleration, yielding zero terminal velocity. Direct position warping can otherwise lead to mesh penetration.
The number of frames used to apply the prismatic joint motion in simulation is formulated as: $N = \left\lceil\frac{\sqrt{2\Delta s_{prismatic}/\mathbf{a}}}{\Delta t} \right\rceil$, where $\mathbf{a}$ is a small constant acceleration, and $\Delta t$ is the unit frame time. At each simulation frame $i$, we apply a fixed force $T_{fixed}$ on the finger tendon if its current displacement $\Delta s^i$ reaches the target teleoperation value $\Delta s_{tendon}$; otherwise, the tendon is released. We formulate frame $i$ tendon force $T_{tendon}^i$ as: 
\begin{align}
T_{tendon}^i =
\begin{cases}
T_{fixed}, & \text{if } \Delta s^i \leq \Delta s_{tendon}, \\[6pt]
0, & \text{otherwise},
\end{cases}
\end{align}


We further augment the teleoperation data in simulation using keyboard control. In the original Nvidia Warp framework, rendering is only available after the simulation ends by exporting each frame to a USD file. To enable real-time interaction, we develop a custom OpenGL-based renderer within Physics Newton~\cite{newton} and integrate it into Warp~\cite{warp2022}, allowing direct teleoperation during simulation. In this setup, the soft-hand teleoperation includes controlling the 6-DoF wrist pose, the prismatic joint, tendon motions, and lift-up motions for each object. Both real-world and simulation teleoperation data are utilized to train the soft hand design. After optimization, we deploy the new design in the real workspace, using teleoperation data augmented in simulation, or directly operated by an operator.
\begin{figure}[thbp]
    \centering
    \includegraphics[width=0.9\linewidth]{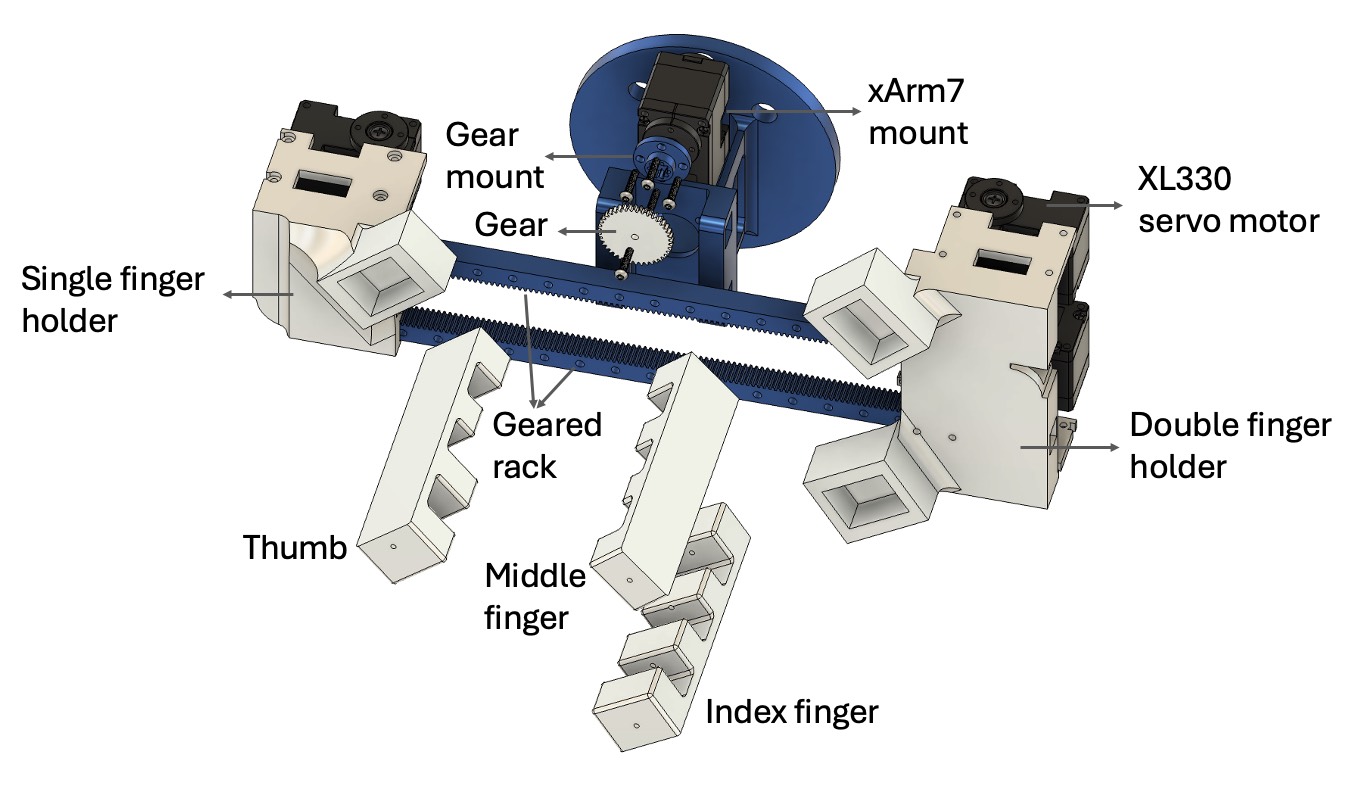}
    \caption{\textbf{Optimized soft robot hand.} We built our final design with optimized fingers, 3D-printed finger holders, an xArm mount, geared racks, and four servo motors.}
    \label{fig:design}
    \vspace{-1.8em}
\end{figure}

\begin{figure*}[htbp]
    \centering
    \begin{subfigure}{0.51\textwidth}
        \centering
        \raisebox{1.0mm}{\includegraphics[width=\linewidth]{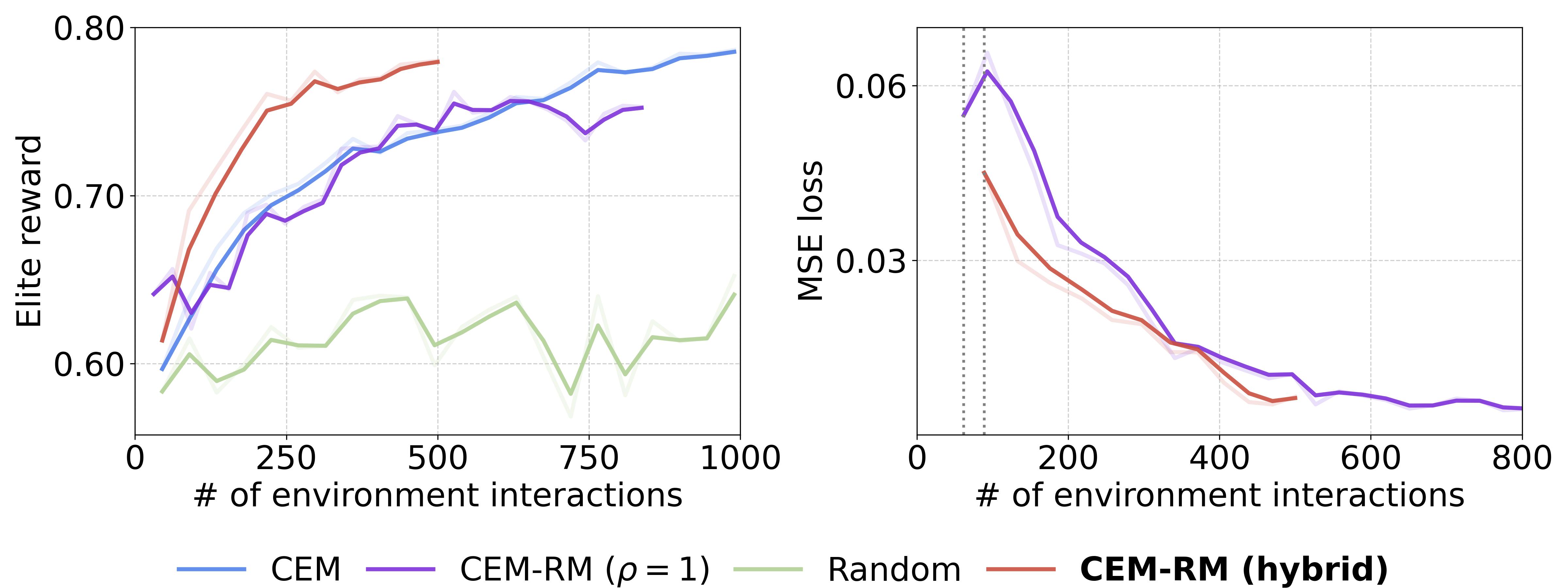}}
        \caption{}
        \label{fig:main}
    \end{subfigure}
    \begin{subfigure}{0.19\textwidth}
        \centering
        \raisebox{-1.0mm}{
        \includegraphics[width=\linewidth]{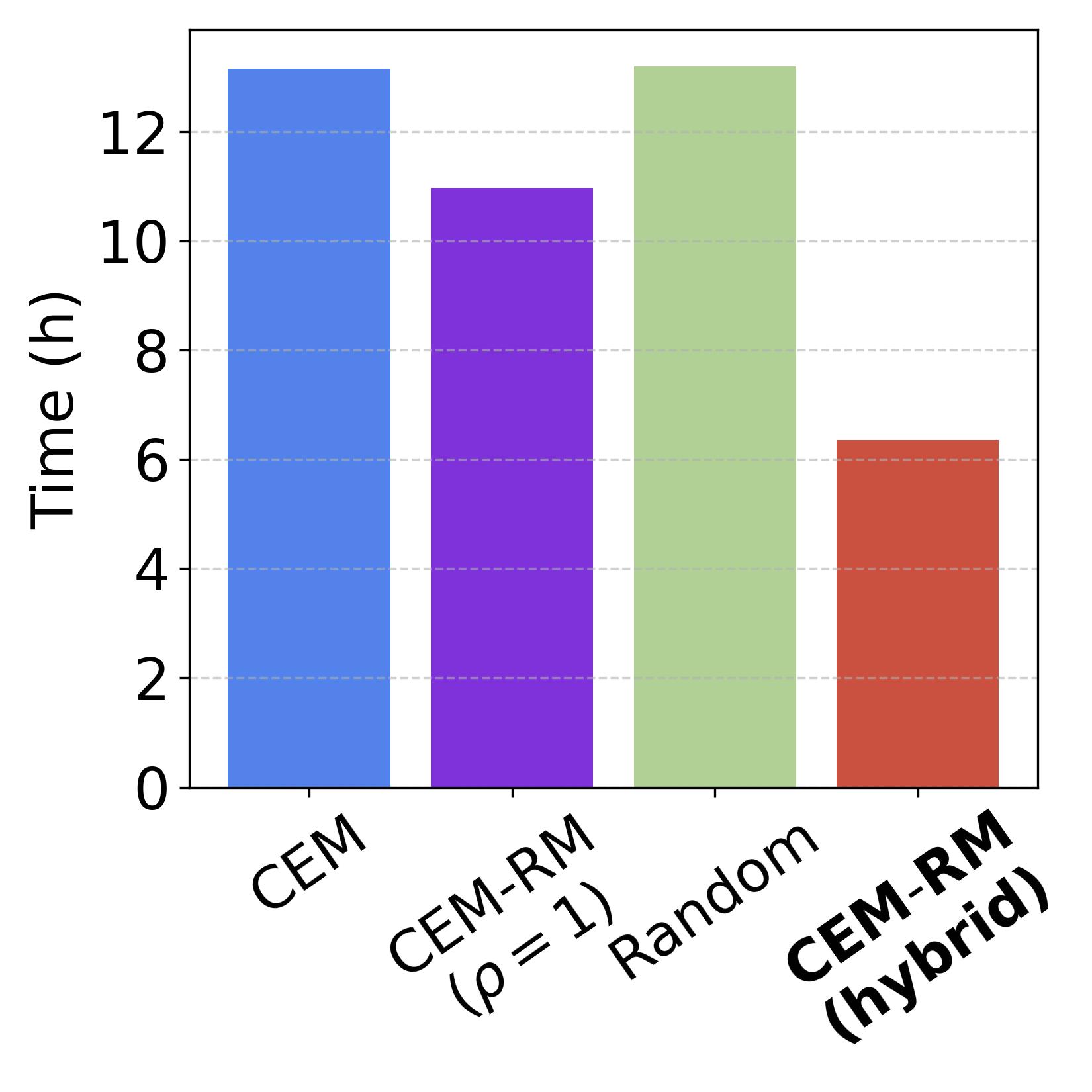}}
        \caption{}
        \label{fig:time}
    \end{subfigure}
    \begin{subfigure}{0.27\textwidth}
        \centering
        \raisebox{2.0mm}{
        \includegraphics[width=\linewidth]{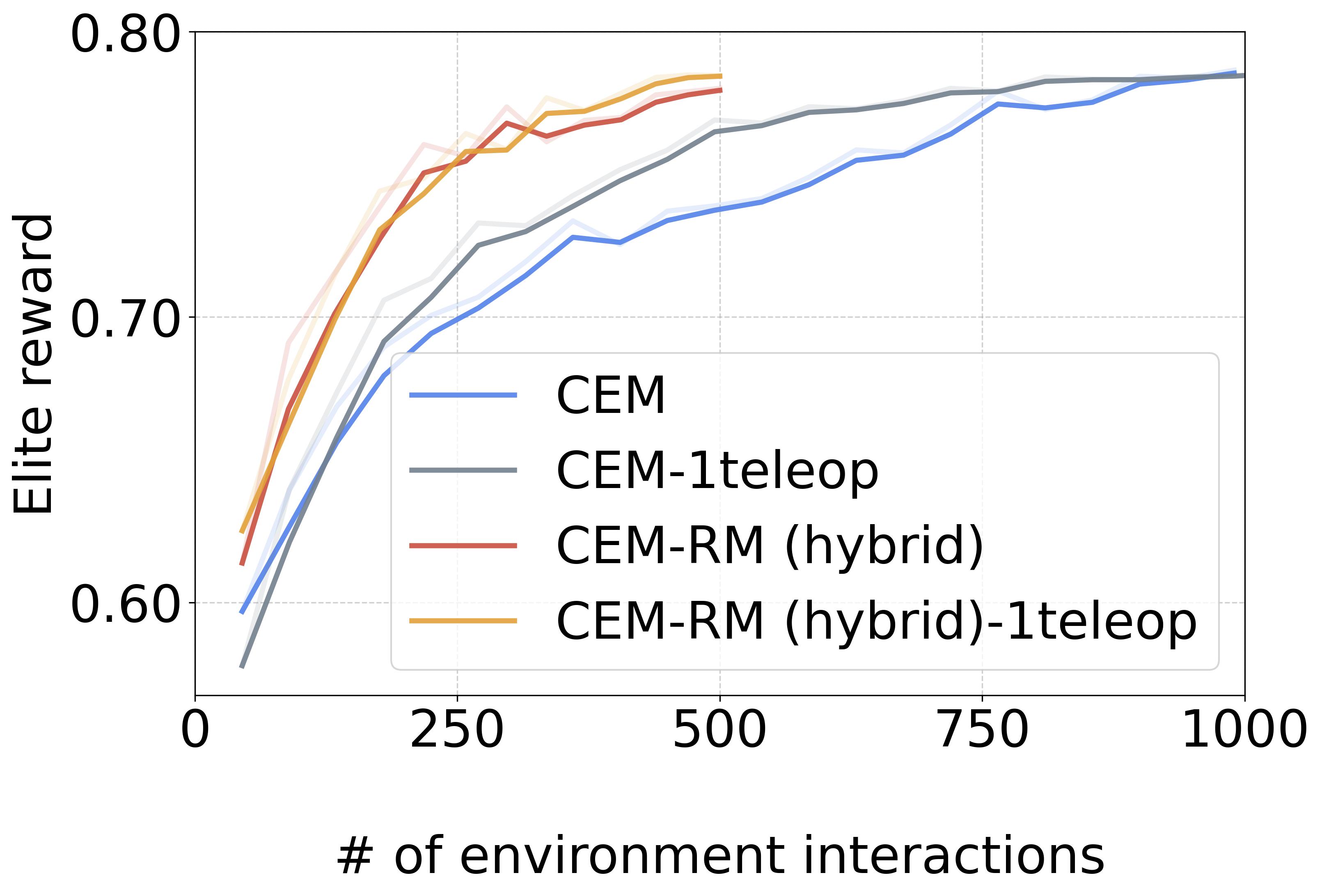}}
        \caption{}
        \label{fig:teleop_abla}
    \end{subfigure}
    \caption{\textbf{Simulation results for optimization methods.}
    (a) Elite reward and loss versus the number of environment interactions, comparing CEM, CEM with reward model, and random sampling. Grey dotted lines indicate the reward-model training start points.
    (b) Convergence training time of the full training process for CEM, CEM-RM ($\rho$=1), Random, and CEM-RM (hybrid).
    (c) Elite reward ablation on teleoperation data: training with single or multiple teleoperation control poses per object.}
    \label{fig:combined}
    \vspace{-1.5em}
\end{figure*}

\section{Experiments and Results}
We conducted a series of experiments in both simulation and the real robotic workspace. The optimized soft robotic hand was evaluated with respect to: (a) the efficiency of our optimization framework, (b) its generalization to diverse objects with varying masses and geometries, and (c) its grasping success rates compared with alternative soft hand designs optimized using other methods.
\subsection{Hardware Setup}
To collect real workspace teleoperation data, we set up a real-world teleoperation system with the xArm7. The operator used Meta Quest 3 to track human hand movements, which were then translated into the arm's joint angles and soft robot hand motion through inverse kinematics (IK) and motion retargeting. We passed through the VR headset in real time for collecting teleoperation data. 

We assembled the optimized soft fingers, finger holders, four Dynamixel XL330 motors, geared racks, and the xArm mount (with slight adjustments to part transforms for clearer presentation in Fig.~\ref{fig:design}). The optimized soft hand design was 3D-printed and mounted onto the xArm, where the motors performed both prismatic joint movements and position-controlled tendon actuation.

For visual perception, we utilized one Intel RealSense L515 camera: positioned at a 30-degree inclination in the front-top workspace region (42.9 cm elevation). We used FoundationPose\cite{wen2024foundationpose} for object's pose detection.

\subsection{Simulation Experiments}
We first used a uniform baseline design ($l_{seg}$ = $l_{fle}$ = 12 mm, $\psi$ = 0, $h_i^{ten}$ = $0.5h_i$ = 11 mm, $\phi_{index}$ = $-\phi_{middle}$ = $-\pi/4$, $\phi_{thumb}$ = 0) to collect five teleoperation data for each object, including grasping poses, prismatic joint values, and tendon pulling forces, both in the real hardware workspace and in simulation. Using this pre-collected teleop-data, we jointly optimized our soft hand design across all objects in the dataset. During optimization, each design candidate was evaluated in parallel on 50 daily and household YCB \cite{YCB} objects over 800 simulation frames. In the test phase, a random downward disturbance impulse between 0.05 and 0.2 N$\cdot$s was applied after grasping, during the subsequent 700 frames at 3000 fps. A grasp was considered successful at the final frame if the object was lifted and securely held by the soft hand: no ground collision occurred, the object maintained contact with the soft hand fingers, and the object’s body exhibited non-zero force feedback. Each training experiment was conducted on a single NVIDIA RTX 4090 GPU, and the corresponding training times are reported in Fig.~\ref{fig:time}.

\textbf{Our CEM with reward model framework is efficient with fewer simulation evaluations.} We leveraged both ground-truth simulation and the reward model to evaluate design candidates during CEM optimization. The evaluation rate $\rho$ for the reward model was gradually increased, as in Eq.~\ref{eq:rho}, with higher reliance placed on the reward model as its fidelity improves during training. We compared this hybrid approach with three baselines: using the reward model exclusively for evaluation ($\rho$=1), pure CEM optimization and random sampling. As shown in Fig.~\ref{fig:main}, our CEM-RM (hybrid) converged faster with fewer environment interactions while achieving a convergence value comparable to that of pure CEM. By combining simulation- and reward-model–based evaluations, the MSE also converged faster compared with relying solely on the reward model. Reward model training began once the buffer size exceeded the batch size.

We further tested the soft hand designs optimized with these four methods on both in-domain and out-of-domain objects, spanning light (density 1–4 $kg/m^2$) and heavy (density 5–8 $kg/m^2$) categories. The out-of-domain set consists of 29 objects drawn from the Digital Twin Catalog (DTC) \cite{DTC}, KIT \cite{KIT}, and YCB datasets (excluding those used in-domain), and features substantially different geometries and sizes compared to the in-domain objects. All evaluations were conducted across the full teleoperation dataset. As shown in Table. \ref{tab:simulation_success}, while all optimized designs performed well when grasping light objects, \textbf{our CEM-RM (hybrid) consistently showed a significant advantage when handling heavier objects.}

We observed several interesting features in the final optimized design. The optimization favors shorter fingers, which improve grasp stability, and flexure blocks tend to be shorter than segment blocks. The fingertips become thicker, offering better frictional and power grip. A lower tendon routing at the fingertip reduces curling, enabling more effective precision grasps of small objects. This aligns with the uniform pressure distribution similar in~\cite{yi2025codesign}. In addition, the thumb is not placed symmetrically opposite the index and middle fingers; instead, it shifts closer to one finger while angling toward the other. This asymmetric placement allows the thumb to collaborate with its neighboring finger, effectively forming a gripper-like pair when actuated in combination with the prismatic joint. Our real experiments suggest that this configuration improves success rates on thin/flat objects where the uniform baseline design often failed due to limited contact areas.
\begin{table}[tbp]
  \centering
   \caption{\textbf{Simulation success rates} for light/heavy in-domain and out-of-domain objects, averaged over 5 trials.}
  \label{tab:simulation_success}
  {\scriptsize                
  \setlength{\tabcolsep}{4pt} 
  \begin{adjustbox}{}
    \begin{tabular}{@{} l c  cc  cc @{}}
      \toprule
      \multicolumn{2}{c}{\textbf{Method}} 
        & \multicolumn{2}{c}{\textbf{In Domain}} 
        & \multicolumn{2}{c}{\textbf{Out of Domain}} \\
      \cmidrule(r){1-2} \cmidrule(r){3-4} \cmidrule(l){5-6}
      \textbf{Group} & \textbf{Multi-teleop} 
        & \textbf{Light} & \textbf{Heavy} 
        & \textbf{Light} & \textbf{Heavy} \\
      
      
      \midrule
      \multirow{1}{*}{CEM-RM($\rho$=1)}&  \cmark   & 86.0\% & 75.2\% & 75.9\%  & 62.1\% \\
      \midrule
      \multirow{1}{*}{Random} 
                                    & \cmark & 86.0\% &  68.0\%&72.4\% &62.1\% \\

      \midrule
      \multirow{2}{*}{CEM}      & \xmark    & 88.4\% & 80.0\% & — & —   \\
      
                                    & \cmark  & 89.2\% & 85.6\% & \textbf{79.3}\% &  65.5\%      \\
    
      \midrule
    \multirow{2}{*}{\textbf{CEM-RM(hybrid)}}
                                    & \xmark  &83.2\% & 82.8\% & — & — \\
                                    & \cmark  & \textbf{89.2\%} & \textbf{87.6}\% &75.9\%  & \textbf{72.4\%} \\
      \bottomrule
    \end{tabular}
  \end{adjustbox}
  }
\vspace{-1.5em}
\end{table}

\begin{figure*}[htpb]
    \centering
    \includegraphics[width=0.9\linewidth]{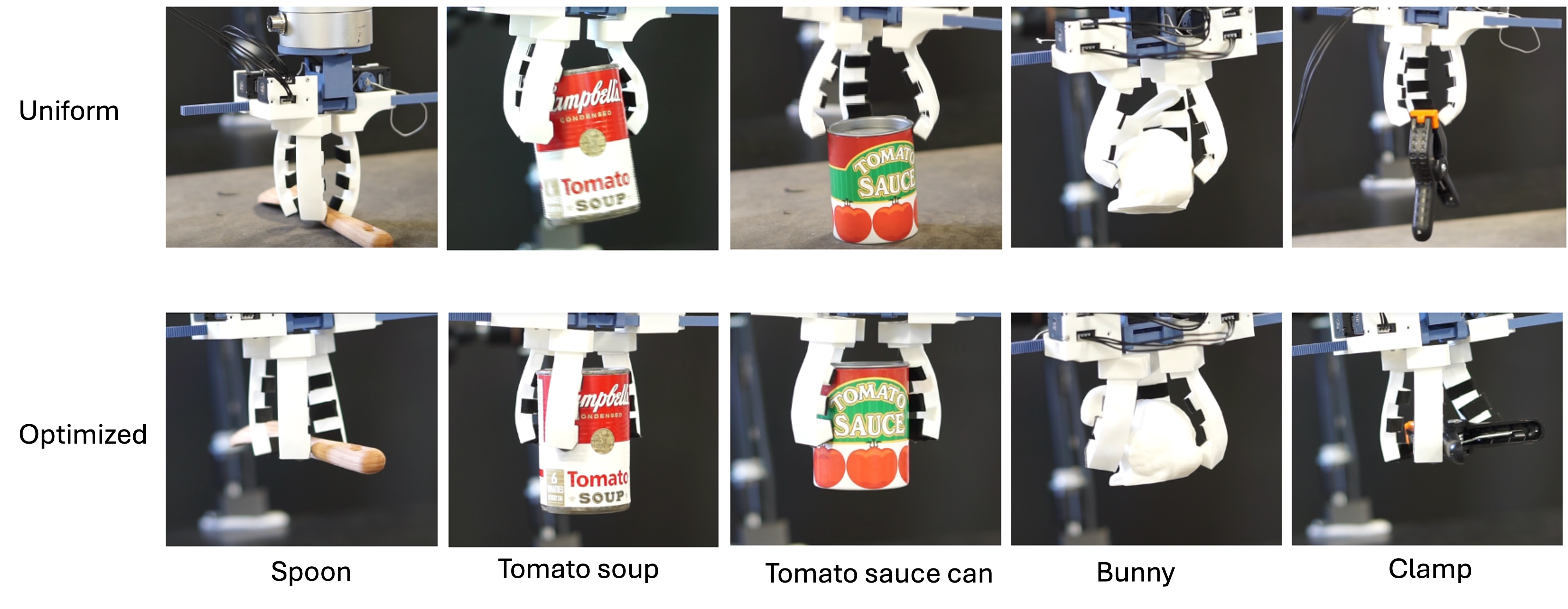}
    \caption{\textbf{Real-world experiments:} comparing uniform and optimized soft hands on in-domain and out-of-domain objects. The uniform hand failed to lift the spoon and tomato sauce can, and could not stably hold the clamp, bunny, and tomato soup, which slipped down after grasping. The optimized design held all objects stably.}
    \label{fig:test}
    \vspace{-1.5em}
\end{figure*}
\subsection{Ablation Study}

We ablated the effect of CEM population size and the use of single or multiple teleoperation control during training. We firstly trained on a small set of 10 objects (randomly selected from the dataset) to explore different population sizes (32/45/56/67), and observed that the elite reward converged to similar values when the size exceeded 45. Thus, we set the population size to 45 for subsequent optimization. To examine the impact of teleop-data diversity, we optimized the soft hand design using either a single teleoperation control pose (we use the first) or all available teleop-data (randomly sampling one per grasp). As shown in Fig.~\ref{fig:teleop_abla}, the single-teleop setting yields slightly higher convergence. However, during the test phase (in Table~\ref{tab:simulation_success}), the single-teleop setting did not achieve higher success rates for either light or heavy objects, indicating that \textbf{training with multiple teleop-data improved robustness.}
\begin{table}[hbp]
  \centering
   \setlength{\tabcolsep}{4pt} 
  \renewcommand{\arraystretch}{1.3}
  \caption{\textbf{Real success rates} for in- and out-of-domain objects of diverse masses and geometries.}
  \begin{adjustbox}{width=0.95\linewidth}
  \begin{tabular}{l*{10}{c}}
 
    \toprule
    Object ID & \textbf{1} & \textbf{2} & \textbf{3} & \textbf{4} &\textbf{5} &\textbf{6} &\textbf{7} &\textbf{8} &\textbf{9}  & \textbf{10} \\
    
    \midrule
    Uniform     & 6/10 & 7/10  & 5/10  & 5/10  &9/10  & 4/10 & 9/10 & 9/10 & 1/10 & 4/10 \\
    Optimized   & 8/10  & 8/10 & 7/10& 7/10 &8/10 & 6/10& 9/10 & 9/10 & 6/10 & 6/10 \\
    Optimized (teleop) & \textbf{8/10} & \textbf{9/10} & \textbf{9/10}  & \textbf{8/10} & \textbf{10/10} & \textbf{8/10} & \textbf{9/10} & \textbf{10/10} & \textbf{9/10} & \textbf{8/10} \\
    \bottomrule
  \end{tabular}
  \end{adjustbox}
  \label{tab:success_rate_real}
  \vspace{-18pt}
\end{table}
\subsection{Real World Experiments}
\begin{figure}[tbp]
    \centering
    \includegraphics[width=0.7\linewidth]{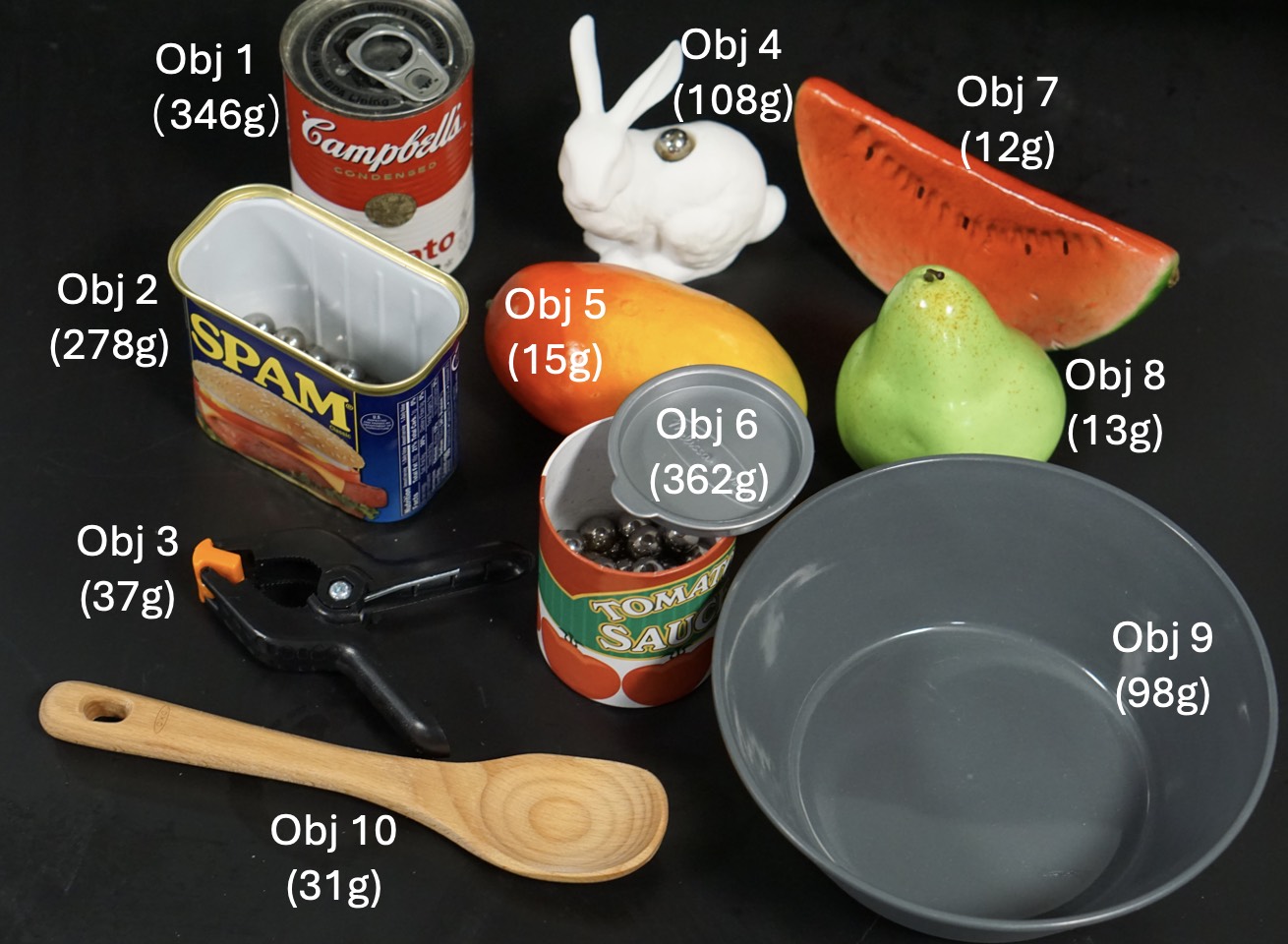}
    \caption{\textbf{Real test objects and their corresponding weights.}}
    \label{fig:obj}
    \vspace{-1.8em}
\end{figure}

We selected 10 objects (Fig.~\ref{fig:obj}) from YCB~\cite{YCB} and the Digital Twin Catalog~\cite{DTC}, covering diverse geometries—flat, thin, cylindrical, irregular, and box-shaped forms—as well as varying masses. We evaluated: 1) the uniform baseline, 2) our optimized design using pre-collected teleoperation data, and 3) real-time teleoperation by an operator to examine whether more precise grasping poses could further enhance performance. As shown in Fig.~\ref{fig:test}, the uniform design struggled to grasp or hold flat objects (spoon) or heavier objects (tomato sauce can) due to its limited compliant contact. When objects were supported only by the fingertips, they eventually slipped to the ground. In contrast, the optimized design—with thicker fingertips and a lower tendon routing at the tip—successfully lifted and held these objects stably. As shown in Table.~\ref{tab:success_rate_real}, real-time teleoperation with the optimized soft hand further improved performance by enabling more precise grasping poses, which are especially beneficial for thin objects (bowl) and heavy objects (tomato sauce can) that demand higher-quality grasp configurations. Overall, the optimized design demonstrates strong performance and robustness across a wide variety of objects.


\section{Conclusion and Limitations}
We developed a broad design space for soft robotic hands, encompassing finger length, tendon routing, and mounting position and orientation, and introduced parallel evaluation in simulation, and proposed a Cross-Entropy Method with Reward Model (CEM-RM) framework to effectively optimize soft hand design using high-quality pre-collected teleoperation data. By incorporating a reward model as a design evaluator, our method enables efficient gradient-based updates while requiring less than half the environment interactions compared to pure optimization approaches. Hardware experiments demonstrate that the resulting soft hand achieved superior grasp robustness on irregularly shaped objects and improved load-bearing capacity relative to both uniform designs and designs optimized without a reward model. Interestingly, the optimized hand exhibits structural features such as thicker fingertips, lower tendon routing at the tips, and a thumb placement that shifts toward one finger while angling toward another. We anticipate that our data-driven optimization framework and the discovered design features will inspire the soft robotics community to develop more creative new designs and sample-efficient learning methods.

\textbf{Limitations.} We currently optimize soft hand design using collected teleoperation control data. However, integrating control policy into the optimization loop in an effective way has the potential to yield better performance. Additionally, more diverse manipulation tasks involving articulated objects can be explored if soft robot–object interactions in simulation remain stable in contact-rich scenarios. A significant sim-to-real gap also persists when transferring teleoperation data for deployment. More precise and robust object detection would help bring soft hand design performance closer to that achieved with real-time teleoperation.

\bibliographystyle{IEEEtran}
\bibliography{ref}

\begin{thebibliography}{10}
\providecommand{\url}[1]{#1}
\csname url@samestyle\endcsname
\providecommand{\newblock}{\relax}
\providecommand{\bibinfo}[2]{#2}
\providecommand{\BIBentrySTDinterwordspacing}{\spaceskip=0pt\relax}
\providecommand{\BIBentryALTinterwordstretchfactor}{4}
\providecommand{\BIBentryALTinterwordspacing}{\spaceskip=\fontdimen2\font plus
\BIBentryALTinterwordstretchfactor\fontdimen3\font minus \fontdimen4\font\relax}
\providecommand{\BIBforeignlanguage}[2]{{%
\expandafter\ifx\csname l@#1\endcsname\relax
\typeout{** WARNING: IEEEtran.bst: No hyphenation pattern has been}%
\typeout{** loaded for the language `#1'. Using the pattern for}%
\typeout{** the default language instead.}%
\else
\language=\csname l@#1\endcsname
\fi
#2}}
\providecommand{\BIBdecl}{\relax}
\BIBdecl

\bibitem{morph}
Z.~He and M.~Ciocarlie, ``Morph: Design co-optimization with reinforcement learning via a differentiable hardware model proxy,'' in \emph{2024 IEEE International Conference on Robotics and Automation (ICRA)}, 2024, pp. 7764--7771.

\bibitem{aloha}
\BIBentryALTinterwordspacing
Z.~Fu, T.~Z. Zhao, and C.~Finn, ``Mobile aloha: Learning bimanual mobile manipulation with low-cost whole-body teleoperation,'' 2024. [Online]. Available: \url{https://arxiv.org/abs/2401.02117}
\BIBentrySTDinterwordspacing

\bibitem{yang2024ace}
S.~Yang, M.~Liu, Y.~Qin, R.~Ding, J.~Li, X.~Cheng, R.~Yang, S.~Yi, and X.~Wang, ``Ace: A cross-platform visual-exoskeletons system for low-cost dexterous teleoperation,'' \emph{arXiv preprint arXiv:2408.11805}, 2024.

\bibitem{khatib2016ocean}
O.~Khatib, X.~Yeh, G.~Brantner, B.~Soe, B.~Kim, S.~Ganguly, H.~Stuart, S.~Wang, M.~Cutkosky, A.~Edsinger \emph{et~al.}, ``Ocean one: A robotic avatar for oceanic discovery,'' \emph{IEEE Robotics \& Automation Magazine}, vol.~23, no.~4, pp. 20--29, 2016.

\bibitem{xin2025analyzingkeyobjectiveshumantorobot}
\BIBentryALTinterwordspacing
C.~Xin, M.~Yu, Y.~Jiang, Z.~Zhang, and X.~Li, ``Analyzing key objectives in human-to-robot retargeting for dexterous manipulation,'' 2025. [Online]. Available: \url{https://arxiv.org/abs/2506.09384}
\BIBentrySTDinterwordspacing

\bibitem{odhner2014compliant}
L.~U. Odhner, L.~P. Jentoft, M.~R. Claffee, N.~Corson, Y.~Tenzer, R.~R. Ma, M.~Buehler, R.~Kohout, R.~D. Howe, and A.~M. Dollar, ``A compliant, underactuated hand for robust manipulation,'' \emph{The International Journal of Robotics Research}, vol.~33, no.~5, pp. 736--752, 2014.

\bibitem{zhai2023desktop}
Y.~Zhai, A.~De~Boer, J.~Yan, B.~Shih, M.~Faber, J.~Speros, R.~Gupta, and M.~T. Tolley, ``Desktop fabrication of monolithic soft robotic devices with embedded fluidic control circuits,'' \emph{Science robotics}, vol.~8, no.~79, p. eadg3792, 2023.

\bibitem{yi2025codesign}
S.~Yi, X.~Bai, A.~Singh, J.~Ye, M.~T. Tolley, and X.~Wang, ``Co-design of soft gripper with neural physics,'' in \emph{Conference on Robot Learning (CoRL)}, 2025.

\bibitem{schulz2017interactive}
A.~Schulz, C.~Sung, A.~Spielberg, W.~Zhao, R.~Cheng, E.~Grinspun, D.~Rus, and W.~Matusik, ``Interactive robogami: An end-to-end system for design of robots with ground locomotion,'' \emph{The International Journal of Robotics Research}, vol.~36, no.~10, pp. 1131--1147, 2017.

\bibitem{liu2018optimal}
C.-H. Liu, T.-L. Chen, C.-H. Chiu, M.-C. Hsu, Y.~Chen, T.-Y. Pai, W.-G. Peng, and Y.-P. Chiang, ``Optimal design of a soft robotic gripper for grasping unknown objects,'' \emph{Soft robotics}, vol.~5, no.~4, pp. 452--465, 2018.

\bibitem{ha2021fit2form}
H.~Ha, S.~Agrawal, and S.~Song, ``Fit2form: 3d generative model for robot gripper form design,'' in \emph{Conference on Robot Learning}.\hskip 1em plus 0.5em minus 0.4em\relax PMLR, 2021, pp. 176--187.

\bibitem{xu2024dynamics}
X.~Xu, H.~Ha, and S.~Song, ``Dynamics-guided diffusion model for robot manipulator design,'' \emph{CoRL}, 2024.

\bibitem{xu2021end}
J.~Xu, T.~Chen, L.~Zlokapa, M.~Foshey, W.~Matusik, S.~Sueda, and P.~Agrawal, ``An end-to-end differentiable framework for contact-aware robot design,'' \emph{arXiv preprint arXiv:2107.07501}, 2021.

\bibitem{kodnongbua2023computational}
M.~Kodnongbua, I.~G.~Y. Lou, J.~Lipton, and A.~Schulz, ``Computational design of passive grippers,'' \emph{arXiv preprint arXiv:2306.03174}, 2023.

\bibitem{chen2018topology}
F.~Chen, W.~Xu, H.~Zhang, Y.~Wang, J.~Cao, M.~Y. Wang, H.~Ren, J.~Zhu, and Y.~Zhang, ``Topology optimized design, fabrication, and characterization of a soft cable-driven gripper,'' \emph{IEEE Robotics and Automation Letters}, vol.~3, no.~3, pp. 2463--2470, 2018.

\bibitem{sliacka2023co}
M.~Sliacka, M.~Mistry, R.~Calandra, V.~Kyrki, and K.~S. Luck, ``Co-imagination of behaviour and morphology of agents,'' in \emph{International Conference on Machine Learning, Optimization, and Data Science}.\hskip 1em plus 0.5em minus 0.4em\relax Springer, 2023, pp. 318--332.

\bibitem{zhao2020robogrammar}
A.~Zhao, J.~Xu, M.~Konakovi{\'c}-Lukovi{\'c}, J.~Hughes, A.~Spielberg, D.~Rus, and W.~Matusik, ``Robogrammar: graph grammar for terrain-optimized robot design,'' \emph{ACM Transactions on Graphics (TOG)}, vol.~39, no.~6, pp. 1--16, 2020.

\bibitem{chen2025physics}
L.~Chen, Y.~Gao, S.~Wang, F.~Fuentes, L.~H. Blumenschein, and Z.~Kingston, ``Physics-grounded differentiable simulation for soft growing robots,'' in \emph{2025 IEEE 8th International Conference on Soft Robotics (RoboSoft)}.\hskip 1em plus 0.5em minus 0.4em\relax IEEE, 2025, pp. 1--8.

\bibitem{georgiev2024pwm}
I.~Georgiev, V.~Giridhar, N.~Hansen, and A.~Garg, ``Pwm: Policy learning with multi-task world models,'' \emph{arXiv preprint arXiv:2407.02466}, 2024.

\bibitem{kulz2024optimizing}
J.~K{\"u}lz and M.~Althoff, ``Optimizing modular robot composition: A lexicographic genetic algorithm approach,'' in \emph{2024 IEEE International Conference on Robotics and Automation (ICRA)}.\hskip 1em plus 0.5em minus 0.4em\relax IEEE, 2024, pp. 16\,752--16\,758.

\bibitem{cma-es}
\BIBentryALTinterwordspacing
N.~Hansen, S.~D. M\"{u}ller, and P.~Koumoutsakos, ``Reducing the time complexity of the derandomized evolution strategy with covariance matrix adaptation (cma-es),'' \emph{Evol. Comput.}, vol.~11, no.~1, p. 1–18, Mar. 2003. [Online]. Available: \url{https://doi.org/10.1162/106365603321828970}
\BIBentrySTDinterwordspacing

\bibitem{ryu2025machine}
J.~Ryu, J.~Kim, C.~W. Park, H.-Y. Kim, D.~Lee, and A.~K. Han, ``Machine learning-based design of 3d-printable microneedles for enhanced tissue anchoring with reduced tissue damage,'' \emph{Advanced Engineering Materials}, p. 2501165, 2025.

\bibitem{softzoo}
T.-H. Wang, P.~Ma, A.~E. Spielberg, Z.~Xian, H.~Zhang, J.~B. Tenenbaum, D.~Rus, and C.~Gan, ``Softzoo: A soft robot co-design benchmark for locomotion in diverse environments,'' \emph{arXiv preprint arXiv:2303.09555}, 2023.

\bibitem{crawlers}
C.~Schaff, A.~Sedal, S.~Ni, and M.~R. Walter, ``Sim-to-real transfer of co-optimized soft robot crawlers,'' \emph{Autonomous Robots}, vol.~47, no.~8, pp. 1195--1211, 2023.

\bibitem{shaw2023leap}
K.~Shaw, A.~Agarwal, and D.~Pathak, ``Leap hand: Low-cost, efficient, and anthropomorphic hand for robot learning,'' \emph{arXiv preprint arXiv:2309.06440}, 2023.

\bibitem{gilday2025embodied}
K.~Gilday, C.~Sirithunge, F.~Iida, and J.~Hughes, ``Embodied manipulation with past and future morphologies through an open parametric hand design,'' \emph{Science Robotics}, vol.~10, no. 102, p. eads6437, 2025.

\bibitem{zorin2025ruka}
A.~Zorin, I.~Guzey, B.~Yan, A.~Iyer, L.~Kondrich, N.~X. Bhattasali, and L.~Pinto, ``Ruka: Rethinking the design of humanoid hands with learning,'' \emph{arXiv preprint arXiv:2504.13165}, 2025.

\bibitem{toshimitsu2023getting}
Y.~Toshimitsu, B.~Forrai, B.~G. Cangan, U.~Steger, M.~Knecht, S.~Weirich, and R.~K. Katzschmann, ``Getting the ball rolling: Learning a dexterous policy for a biomimetic tendon-driven hand with rolling contact joints,'' in \emph{2023 IEEE-RAS 22nd International Conference on Humanoid Robots (Humanoids)}.\hskip 1em plus 0.5em minus 0.4em\relax IEEE, 2023, pp. 1--7.

\bibitem{kim2025exo}
B.~Kim, U.~Jeong, and K.-J. Cho, ``Exo-glove pinch: A soft, hand-wearable robot designed through constrained tendon routing analysis,'' \emph{IEEE Robotics and Automation Letters}, 2025.

\bibitem{mohammadi2020practical}
A.~Mohammadi, J.~Lavranos, H.~Zhou, R.~Mutlu, G.~Alici, Y.~Tan, P.~Choong, and D.~Oetomo, ``A practical 3d-printed soft robotic prosthetic hand with multi-articulating capabilities,'' \emph{PloS one}, vol.~15, no.~5, p. e0232766, 2020.

\bibitem{kim2019continuously}
J.~Kim, W.-Y. Choi, S.~Kang, C.~Kim, and K.-J. Cho, ``Continuously variable stiffness mechanism using nonuniform patterns on coaxial tubes for continuum microsurgical robot,'' \emph{IEEE Transactions on Robotics}, vol.~35, no.~6, pp. 1475--1487, 2019.

\bibitem{mnih2013playing}
V.~Mnih, K.~Kavukcuoglu, D.~Silver, A.~Graves, I.~Antonoglou, D.~Wierstra, and M.~Riedmiller, ``Playing atari with deep reinforcement learning,'' \emph{Nature 518}, p. 529–533, 2015.

\bibitem{Lillicrap2015ContinuousCW}
\BIBentryALTinterwordspacing
T.~P. Lillicrap, J.~J. Hunt, A.~Pritzel, N.~M.~O. Heess, T.~Erez, Y.~Tassa, D.~Silver, and D.~Wierstra, ``Continuous control with deep reinforcement learning,'' \emph{CoRR}, vol. abs/1509.02971, 2015. [Online]. Available: \url{https://api.semanticscholar.org/CorpusID:16326763}
\BIBentrySTDinterwordspacing

\bibitem{schulman2017proximal}
J.~Schulman, F.~Wolski, P.~Dhariwal, A.~Radford, and O.~Klimov, ``Proximal policy optimization algorithms,'' \emph{arXiv preprint arXiv:1707.06347}, 2017.

\bibitem{Haarnoja2018SoftAA}
T.~Haarnoja, A.~Zhou, K.~Hartikainen, G.~Tucker, S.~Ha, J.~Tan, V.~Kumar, H.~Zhu, A.~Gupta, P.~Abbeel, and S.~Levine, ``Soft actor-critic algorithms and applications,'' \emph{ArXiv}, vol. abs/1812.05905, 2018.

\bibitem{RUBINSTEIN199789}
R.~Y. Rubinstein, ``Optimization of computer simulation models with rare events,'' \emph{European Journal of Operational Research}, vol.~99, no.~1, pp. 89--112, 1997.

\bibitem{kalashnikov2018scalable}
D.~Kalashnikov, A.~Irpan, P.~Pastor, J.~Ibarz, A.~Herzog, E.~Jang, D.~Quillen, E.~Holly, M.~Kalakrishnan, V.~Vanhoucke \emph{et~al.}, ``Scalable deep reinforcement learning for vision-based robotic manipulation,'' in \emph{Conference on robot learning}.\hskip 1em plus 0.5em minus 0.4em\relax PMLR, 2018, pp. 651--673.

\bibitem{lowrey2018plan}
\BIBentryALTinterwordspacing
K.~Lowrey, A.~Rajeswaran, S.~Kakade, E.~Todorov, and I.~Mordatch, ``Plan online, learn offline: Efficient learning and exploration via model-based control,'' in \emph{International Conference on Learning Representations}, 2019. [Online]. Available: \url{https://openreview.net/forum?id=Byey7n05FQ}
\BIBentrySTDinterwordspacing

\bibitem{Bhardwaj2021BlendingM}
M.~Bhardwaj, S.~Choudhury, and B.~Boots, ``Blending mpc \& value function approximation for efficient reinforcement learning,'' \emph{ArXiv}, vol. abs/2012.05909, 2021.

\bibitem{sikchi2022learning}
H.~Sikchi, W.~Zhou, and D.~Held, ``Learning off-policy with online planning,'' in \emph{Conference on Robot Learning}.\hskip 1em plus 0.5em minus 0.4em\relax PMLR, 2022, pp. 1622--1633.

\bibitem{Hansen2022tdmpc}
N.~Hansen, X.~Wang, and H.~Su, ``Temporal difference learning for model predictive control,'' in \emph{International Conference on Machine Learning (ICML)}, 2022.

\bibitem{williams1992Reinforce}
R.~J. Williams, ``Simple statistical gradient-following algorithms for connectionist reinforcement learning,'' \emph{Machine Learning}, vol.~8, p. 229–256, 1992.

\bibitem{guo2025deepseek}
D.~Guo, D.~Yang, H.~Zhang, J.~Song, R.~Zhang, R.~Xu, Q.~Zhu, S.~Ma, P.~Wang, X.~Bi \emph{et~al.}, ``Deepseek-r1: Incentivizing reasoning capability in llms via reinforcement learning,'' \emph{arXiv preprint arXiv:2501.12948}, 2025.

\bibitem{warp2022}
M.~Macklin, ``Warp: A high-performance python framework for gpu simulation and graphics,'' \url{https://github.com/nvidia/warp}, March 2022, nVIDIA GPU Technology Conference (GTC).

\bibitem{newton}
\BIBentryALTinterwordspacing
{Newton Contributors}, ``{Newton}: {GPU}-accelerated physics simulation for robotics, and simulation research.'' {Newton a Series of LF Projects, LLC}, 2025. [Online]. Available: \url{https://github.com/newton-physics/newton}
\BIBentrySTDinterwordspacing

\bibitem{wen2024foundationpose}
B.~Wen, W.~Yang, J.~Kautz, and S.~Birchfield, ``Foundationpose: Unified 6d pose estimation and tracking of novel objects,'' in \emph{Proceedings of the IEEE/CVF Conference on Computer Vision and Pattern Recognition}, 2024, pp. 17\,868--17\,879.

\bibitem{YCB}
\BIBentryALTinterwordspacing
B.~{\c{C}}alli, A.~Walsman, A.~Singh, S.~S. Srinivasa, P.~Abbeel, and A.~M. Dollar, ``Benchmarking in manipulation research: The {YCB} object and model set and benchmarking protocols,'' \emph{CoRR}, vol. abs/1502.03143, 2015. [Online]. Available: \url{http://arxiv.org/abs/1502.03143}
\BIBentrySTDinterwordspacing

\bibitem{DTC}
\BIBentryALTinterwordspacing
Z.~Dong, K.~Chen, Z.~Lv, H.-X. Yu, Y.~Zhang, C.~Zhang, Y.~Zhu, S.~Tian, Z.~Li, G.~Moffatt, S.~Christofferson, J.~Fort, X.~Pan, M.~Yan, J.~Wu, C.~Y. Ren, and R.~Newcombe, ``Digital twin catalog: A large-scale photorealistic 3d object digital twin dataset,'' 2025. [Online]. Available: \url{https://arxiv.org/abs/2504.08541}
\BIBentrySTDinterwordspacing

\bibitem{KIT}
\BIBentryALTinterwordspacing
A.~Kasper, Z.~Xue, and R.~Dillmann, ``The {KIT} object models database: An object model database for object recognition, localization and manipulation in service robotics,'' \emph{Int. J. Robotics Res.}, vol.~31, no.~8, pp. 927--934, 2012. [Online]. Available: \url{https://doi.org/10.1177/0278364912445831}
\BIBentrySTDinterwordspacing

\end{thebibliography}

\end{document}